# Translation Entropy: A Statistical Framework for Evaluating Translation Systems

**Ronit D. Gross[a,1], Yanir Harel[a,1] and Ido Kanter[a,b,*]**

[a]Department of Physics, Bar-Ilan University, Ramat-Gan, 52900, Israel.

[b]Gonda Interdisciplinary Brain Research Center, Bar-Ilan University, Ramat-Gan, 52900, Israel.

*Corresponding author at: Department of Physics, Bar-Ilan University, Ramat-Gan, 52900, Israel. E-mail address: ido.kanter@biu.ac.il (I. Kanter).

[1]These authors equally contributed to this work

## Abstract

The translation of written language has been known since the 3rd century BC; however, its necessity has become increasingly common in the information age. Today, many translators exist, based on encoder-decoder deep architectures, nevertheless, no quantitative objective methods are available to assess their performance, likely because the entropy of even a single language remains unknown. This study presents a quantitative method for estimating translation entropy, with the following key finding. Given a translator, several sentences that differ by only one selected token of a given pivot sentence yield identical translations. Analyzing the statistics of this phenomenon across an ensemble of such sentences, consisting each of a pivot selected token, yields the probabilities of replacing this specific token with others while preserving the translation. These probabilities constitute the entropy of the selected token, and the average across all selected pivot tokens provides an estimate of the translator's overall translation entropy, which is enhanced along the decoder blocks. This entropic measure allows for the quantitative ranking of several publicly available translators and reveals whether mutual translation entropy is symmetric. Extending the proposed method to include the replacement of two tokens in a given pivot sentence demonstrates a multiplicative effect, where translation degeneracy is proportional to the product of the degeneracies of the two tokens. These findings establish translation entropy as a measurable property and objective benchmarking of artificial translators. Results are based on MarianMT, T5-Base and NLLB-200 translators.

## 1. Introduction

The earliest known writing systems, the Sumerian archaic cuneiform script and Egyptian hieroglyphs, date back to the late 4th millennium BC[1-3]. However, the earliest known literature, the Epic of Gilgamesh, dates to 2,100 BC and consists of five Sumerian poems about Gilgamesh, the king of Uruk[4].

As modes of transportation improved and populations expanded, humans became less localized. This led to more frequent interactions among diverse groups speaking different languages, creating the need for translation. Probably the first known translation, the Septuagint, also called the Greek Old Testament or the Translation of the Seventy, was completed around 250 BC[5-7]. Commissioned by Ptolemy II Philadelphus, the original Biblical Hebrew text was translated into Greek by 72 Hebrew translators working independently, yet all produced identical translations. However, translation can also involve interpretation, which may lead to variations in meaning. An early example of such translation is Targum (Translation) Onkelos from the early second century CE[8, 9], the primary Jewish Aramaic translation of the original Biblical Hebrew. Over the past two millennia, many translations have been documented across various fields and languages, now numbering over a thousand, with their exact number depending on the dichotomy between languages and dialects.

Unlike standardized systems of units such as the Meter-Kg-Second (MKS) system, which became a semi-universal system, or the Euro, which became the official currency of about 20 countries, the number of spoken languages remains vast. Attempts to introduce a simple, universal spoken language, such as Esperanto, have largely been unsuccessful[10]. Nevertheless, in recent decades, a few universal programmable languages that operate independently of native keyboard languages have emerged.

Recent advances in large language models (LLMs), based on deep learning architectures[11-13], have enabled the development of sophisticated translation systems. These translators use encoder-decoder architectures[14, 15] trained to convert text from one language to another. Fine-tuning a single trained model allows it to translate among multiple languages, suggesting that such models possess underlying universal structures.

The mathematical study of languages dates back to antiquity; however, the present work focuses on the open question of measuring the entropy of a language. This concept is closely related to the optimal compression of text and the redundancy required to ensure reliable communication over noisy channels. The goal is to estimate the number of grammatically valid and meaningful texts in a given language—English, for example—composed of $N$ letters. As English contains $27$ letters (including the space), the theoretical upper bound for possible combinations is $27^N$. However, the actual number is far smaller due to grammatical and phonetical constraints. Shannon performed early estimations using $N = 2$ and $N = 3$ [16]; however, extending this analysis to larger $N$ was computationally infeasible at the time. Even today, estimating the entropy for $N > 10$ remains impossible for two main reasons: first, computational complexity increases exponentially with $N$; and second, accurately estimating rare events would require datasets larger than $27^N$, which are unachievable. Aware of these limitations, Shannon devised an experimental method in which participants were shown a portion of text and asked to predict subsequent letters[16]. From these experiments, Shannon estimated that the number of meaningful and grammatically valid texts grows as $\sim 2^N$, yielding an entropy of about one bit per letter—an estimate that still holds.

The theoretical estimation of language entropy remains unresolved[17-19]; thus, estimating translation entropy (TE) (the entropy between two languages) seems an even more challenging theoretical problem. Nevertheless, this study introduces a method to estimate TE without requiring the entropy of each language to be evaluated.

The main finding is that, given an encoder-decoder translator, several sentences that differ by only one selected token from a given pivot sentence yield identical translations. Analyzing this phenomenon across an ensemble of sentences, each containing one of the selected tokens, results in the probabilities to replace this token by other ones, while preserving the translation. These probabilities constitute the entropy of the selected token, and their average across all selected pivot tokens provides an estimate of the TE of the translator, which is enhanced along the decoder blocks. This entropic measure enables the quantitative ranking of several publicly available translators and also reveals whether the mutual TE between two languages is symmetric. Finally, extending the proposed method to include the replacement of two tokens in each pivot sentence demonstrates a multiplicative phenomenon, where translation degeneracy is proportional to the product of the degeneracies of the two tokens.

## 2. Results

### 2.1. Formulation of translation entropy exemplified using MarianMT translator

The TE estimation method is first exemplified using the MarianMT (Helsinki-NLP/opus-mt) deep learning architecture, consisting of six encoder blocks followed by six decoder blocks[20]. The model was trained on the Opus100 dataset, which contains one million training sentences and $2{,}000$ validation sentences[21, 22], each provided in both source and target (translated) versions[23]. Results are presented for translations from English (source) to French (target), where each language comprises approximately $30{,}000$ tokens (Fig. 1a). Subsequent results are presented for the reverse translation (from French to English).

In the first step of the algorithm, a token $T^1$ is selected. Next, a pivot English source sentence containing up to $128$ tokens, $T_1, \dots, T_{128}$, is chosen such that $T_j = T^1$ at position $j$ (Fig. 1a, top). The model then generates the corresponding French translation of this pivot sentence (Fig. 1a, top). In the second step, $T^1$ at position $j$ is sequentially replaced with all other possible source tokens, producing $\sim 30{,}000$ generated French sentences (Fig. 1a). Generated sentences that are identical to the pivot-generated translation are identified, as presented in Fig. 1a and schematically illustrated in Fig. 1b.

This form of translation degeneracy, where several source sentences differ from the pivot only by the selected token yet produce identical translations, is exemplified in Fig. 2. This conditional measure differs from cases where the selected pivot token can be replaced within a specific sentence while still yielding the same translation.

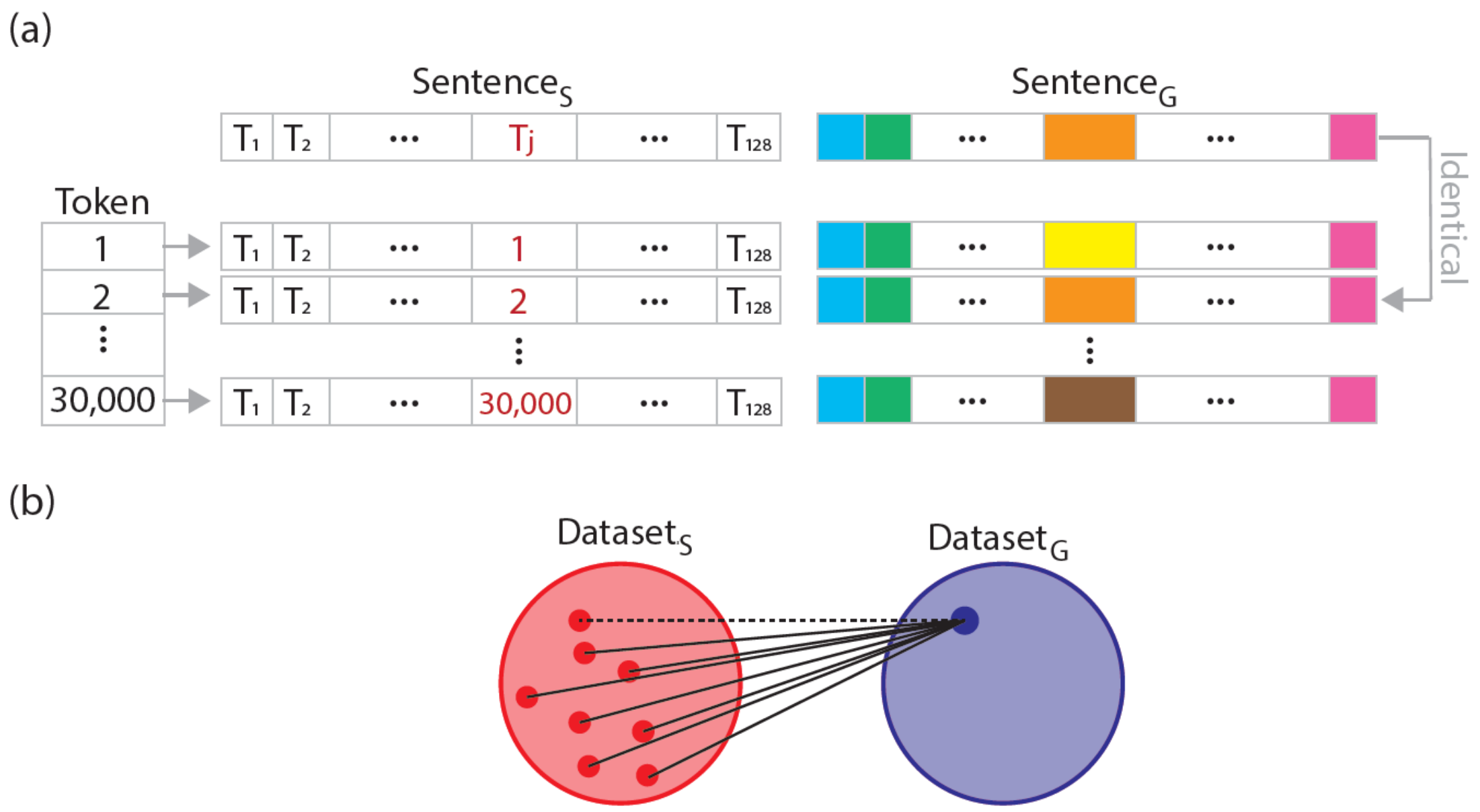

**Fig. 1.** Schematic of translation degeneracy. (a) A pivot English source sentence, $Sentence_S$, consisting of up to 128 tokens, $T_1, \dots, T_{128}$, including $T_j = T^1$ at position $j$, and its generated French sentence, $Sentence_G$ (color-coded) (top). $T^1$ is sequentially replaced by $\sim$30,000 other possible source tokens, resulting in corresponding French-generated sentences (color-coded). Identical generated sentences to the pivot-generated sentence are indicated. (b) Schematic representation of panel a. The dashed line represents a pivot-source sentence (red circle) connected to its generated sentence (blue circle). Other red circles represent sentences differing from the pivot by token $T^1$ at position $j$ and yielding identical generated sentences (blue circles). Note that $Dataset_S$ contains all possible source sentences.

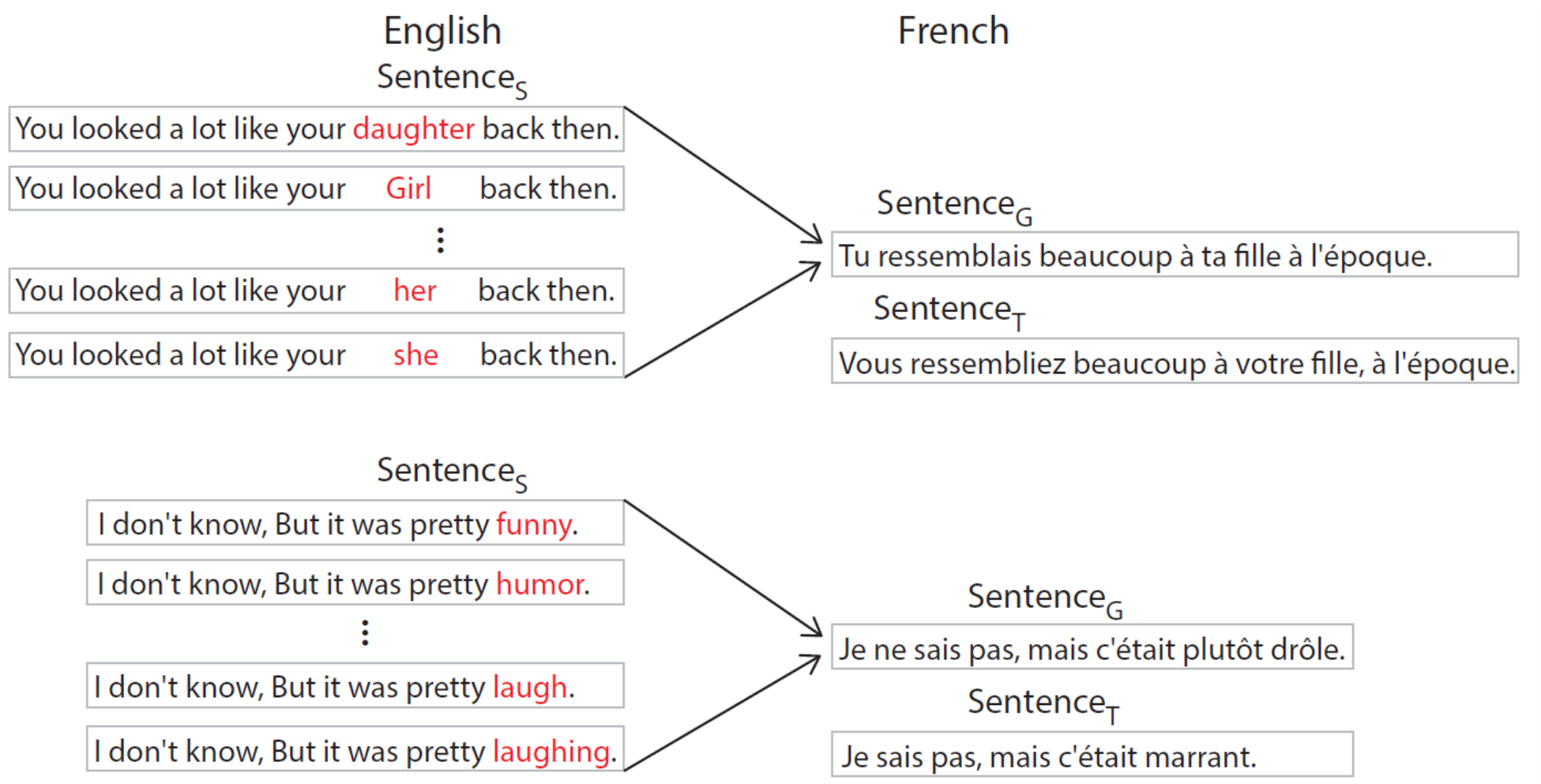

**Fig. 2.** Examples of translation degeneracy. A selected token “daughter” (red) in an English source pivot sentence, its generated and target (true) sentences in French, and three additional source sentences (differing in the red token) that yield an identical generated sentence (Top). A similar example is shown for the selected token “funny” in the pivot source sentence (bottom).

In the second step of the algorithm, the first step was repeated over $30$ pivot source sentences, each containing the same selected token $T^1$ (Fig. 1a, top and Fig. 2). Each of these sentences, $m = 1,\ldots,30$, generates a subgroup of tokens, $SG_m(T^1)$ that preserve the pivot-generated translation. The average subgroup size is given by $|SG(T^1)| =< |SG_m(T^1)| >_m$. Notably, the size of $SG_m(T^1)$ can exceed several thousands for certain irregular source sentences in which the model ignores $T^1$ in the generated translation. In such cases, nearly any token becomes a suitable replacement. For example, in the source sentence “(gunshots, dog groaning) weren't you supposed to be getting some wine?”, the generated sentence “Tu n'étais pas censé prendre du vin ?” omits the word “dog”. Consequently, the subgroup size $|SG(T^1)|$ for the token representing “dog” is $\sim 29{,}000$. To mitigate the influence of such meaningless degeneracies—which can disproportionately inflate $|SG(T^1)|$—only the $24$ smallest subgroup sizes $|SG_m(T^1)|$ were considered, denoted as $SG^{24}(T^1)$. However, the qualitative results were found to be insensitive to the specific threshold chosen.

The number of appearances of each token in $SG^{24}(T^1)$ yields the number of times, between $1$ and $24$, that $T^1$ can be replaced by another token across the $24$ sentences while preserving the generated translation (Fig. 3). Dividing these counts by $24$ produces the probabilities, $P_i$, representing the likelihood that token $i$ can replace $T^1$ while maintaining the same translation. These probabilities are valid under the assumption of sufficiently large number of source sentences, containing $T^1$, and their uniformity leads to two quantities: the average number of times $N_{Av}(T^1)$, $T^1$ can be replaced by another token belonging to $SG^{24}(T^1)$ (Fig. 3)

$$N_{Av}(T^1) = 24 \sum_i P_i \quad (1)$$

and the entropy[24],

$$S(T^1) = -\sum_i P_i \log_2 P_i \quad (2) \quad .$$

As the summation of the probabilities, $\sum P_i$, is not normalized, the entropy, $S(T^1)$, is not simply bounded and depends on the distribution profile of $P_i$. For instance, if there exists an additional token (besides the pivot token) with $P_i = 1$, it does not contribute to the entropy, as it most likely represents a strong synonym—similarly the contribution to the entropy is minimal for a value approaching $P_i \to 1$. As $P_i$ decreases, the token becomes a weaker match to the pivot token, increasing its entropy contribution until reaching a maximum at $P_i = \frac{1}{e} \sim 0.37$ $(N_{AV} \sim 9)$[25]. For $P_i < 0.37$, the token represents noise—a mismatched substitution for the pivot token— whose entropy contribution decreases with $P_i$. This behavior is expected for large sample sizes $(m \to \infty)$; however, for smaller sample size examined here $(m = 24)$, larger fluctuations are expected.

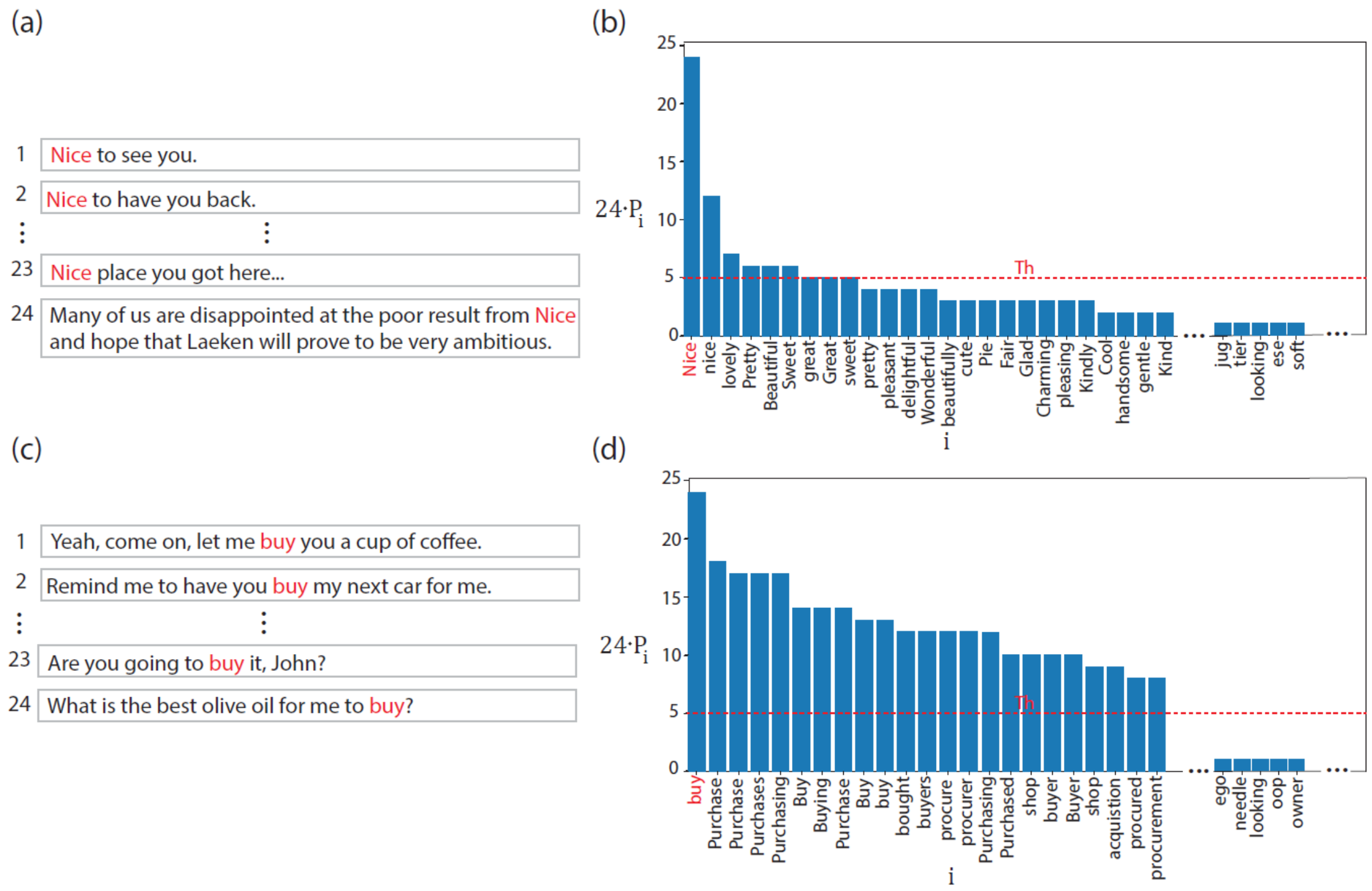

**Fig. 3.** (a) Examples of 24 selected pivot source sentences containing the token “Nice” (red). (b) Corresponding $24 \cdot P_i$ values constituting $N_{Av}(T^1)$ (Eq. (1)), obtained from the 24 selected pivot sentences in panel a. $24 \cdot P_i$ indicates the number of times the token “Nice” can be replaced by token $i$ while preserving the generated translation. The horizontal dashed red line marks the selected threshold. Many tokens with $24 \cdot P_i = 1$ are not shown. (c-d) Same as panels a-b but for the token “buy.”

The pivot token can be replaced, with a non-negligible probability, by semantically related tokens while preserving the generated pivot translation. For example, the token “Nice” can be replaced by the tokens “nice” or “lovey” with relatively high probabilities while preserving the translation (Fig. 3). However, numerous meaningless tokens, such as “jug,” “broad,” or “ese,” can also replace “Nice” with very low probabilities (Fig. 3b). These meaningless tokens arise because the translator occasionally substitutes such gibberish tokens with the correct token “Nice” at low probabilities, approximately $\frac{\beta}{24}$ and $\beta \to 1$ (Fig. 3b). The main entropy contribution from such gibberish tokens to the entropy is $\sim \frac{\beta}{24} log_2(24)$ (Eq. (2)). Their cumulative entropy is on the order of $O(10^3)$, which dominates the average entropy because their number can exceed several thousand for certain pivot tokens (Fig. 4a). In contrast, meaningful replacement tokens have higher $\beta$ values. Therefore, the entropy per pivot token (Eq. (2)) was calculated using the following threshold criterion:

$$P_i > Threshold = \frac{\beta_c}{24} \quad (4)$$

where presented results are for $\beta_c = 5$ (Fig. 4b), however the qualitative results and conclusions remain robust within the range $5 \leq \beta_c \leq 10$.

Statistical characteristics of the translating process were obtained by repeating the first two steps over $100$ randomly selected pivot source tokens, $T^\alpha$ ($\alpha = 1, \dots, 100$). The quality of the MarianMT (Helsinki-NLP/opus-mt) translator for English-to-French translation was quantified using the following averaged entropy:

$$S =< S(T^\alpha) >_\alpha \qquad (5)$$

where $< \cdots >_\alpha$ denotes the average over the selected pivot source tokens, ordered by increasing entropy (Fig. 4a). The results indicate large fluctuations—spanning three orders of magnitude—among different $T^\alpha$ (Fig. 4a). When applying the threshold (Eq. (3)), almost all entropies, $S(T^\alpha)$ fall within the range of $\sim(1, 13)$, although a small number of tokens still exhibit unusually high value (Fig. 4b). These residual high entropies $S(T^\alpha)$ likely result from incomplete filtering of meaningless tokens within $SG_{24}(T^\alpha)$, some of which still satisfy $P_i > Threshold$ (Eq. (4)) for a large number of tokens. To minimize the effects of such rare outliers, the average entropy was recalculated using only the 95 lowest $S(T^{\alpha})$ values, yielding $S^{95} \sim 3.8$ (Fig. 4c).

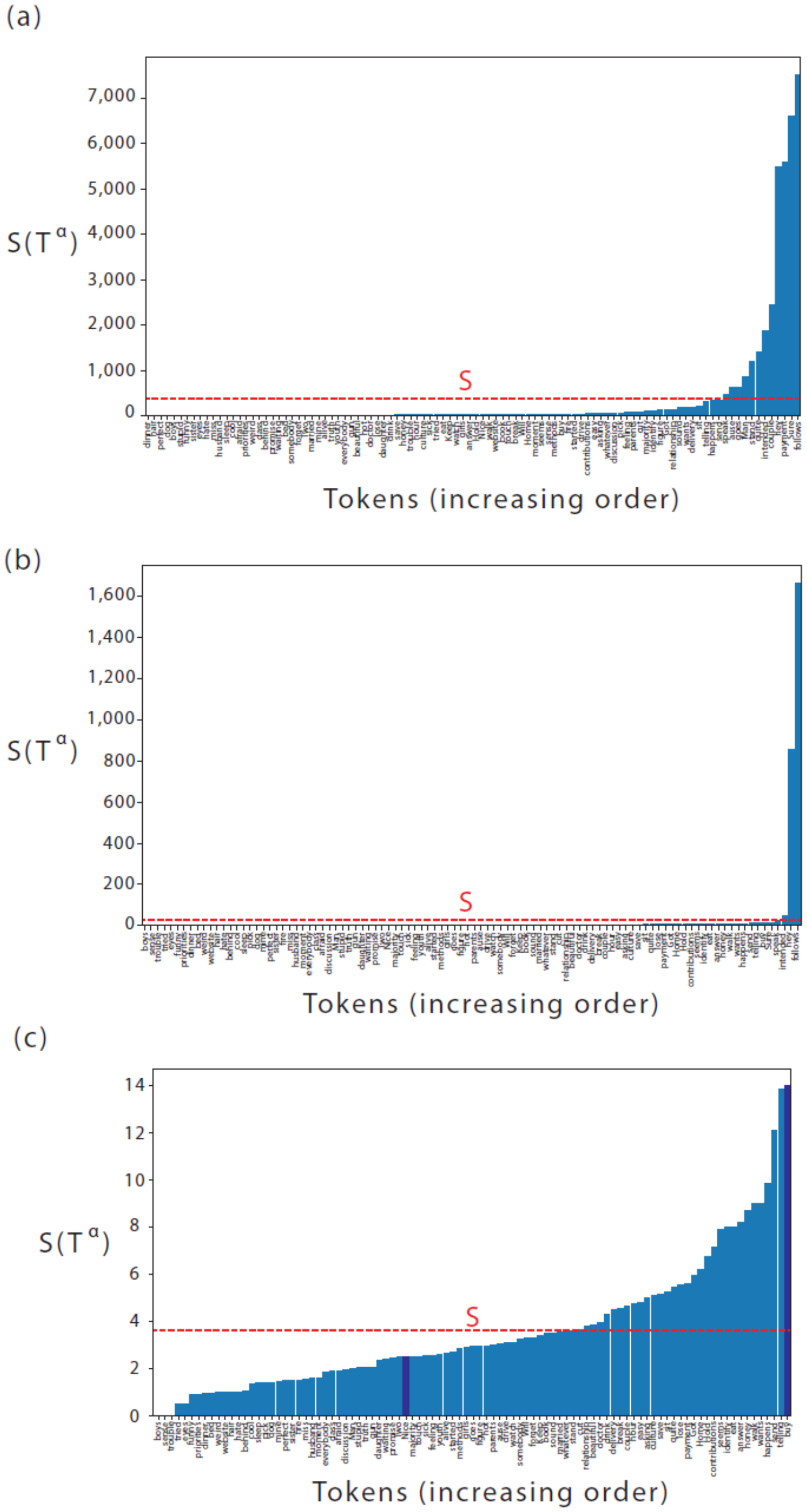

**Fig. 4.** (a) Entropy values for 100 randomly selected pivot source tokens, presented in increasing rank order using MarianMT(Helsinki-NLP/opus-mt) translator, with the average entropy S indicated (horizontal dashed red line). (b) Similar to panel a, but with entropies calculated using $Threshold = \frac{5}{24}$ (Eq. (4)).

(c) Similar to panel b, but showing the $95$ lowest $S(T^{\alpha})$. The average entropy in each panel is indicated (horizontal dashed red line).

An example of a pivot token with relatively low entropy, $S{\sim}2$, is “Nice” (left dark blue bin in Fig. 4c), which has only a few probabilities $P_i > Threshold$ (Fig. 3b). In contrast, the pivot token “buy” exhibits a higher entropy, $S{\sim}14$ (right dark blue bin in Fig. 4c), as many of its replacement probabilities exceed the threshold $(P_i > Threshold)$ (Fig. 3d).

The $30$ pivot source sentences were selected from the Opus100 dataset, which served as the training dataset for MarianMT (Helsinki-NLP/opus-mt). The Opus100 test dataset contained only $2{,}000$ sentences; thus, obtaining $30$ pivot sentences for each of the $100$ pivot tokens is impossible. Nevertheless, training MarianMT on only $200{,}000$ sentences from Opus100 yielded the same entropy values for the $30$ pivot sentences selected from this smaller subset as for those selected from the remaining $800{,}000$ sentences.

Pivot source tokens were chosen to appear between $500$ and $1{,}500$ times in the training dataset. The upper limit excluded highly frequent tokens, such as conjunctions, while the lower limit excluded rare tokens and ensured the validity of the entropy approximation (Eq. (2)). A more accurate estimation of $N_{Av}$ and $S$ could be achieved by increasing the number of examined source tokens beyond $100$ and the token frequency range beyond $[500, 1{,}500]$.

### 2.2. Translation from French to English using MarianMT translator

The same TE estimation procedure was applied for French-to-English translation using the trained encoder-decoder MarianMT (Helsinki-NLP/opus-mt) model. The $100$ pivot source tokens in French were selected similarly to the selection method used in English.

Entropy fluctuations among the selected French source tokens, $S(T^{\alpha})$ (Eq. (2)), were much larger than those observed in the reverse English-to-French translation (Figs. 4 and 5). Moreover, the average entropy was significantly higher in the French-to-English translation (Table 1). For instance, the average entropy increased by $\sim 2.5$ times when comparing the $S^{95}$ values (Table 1).

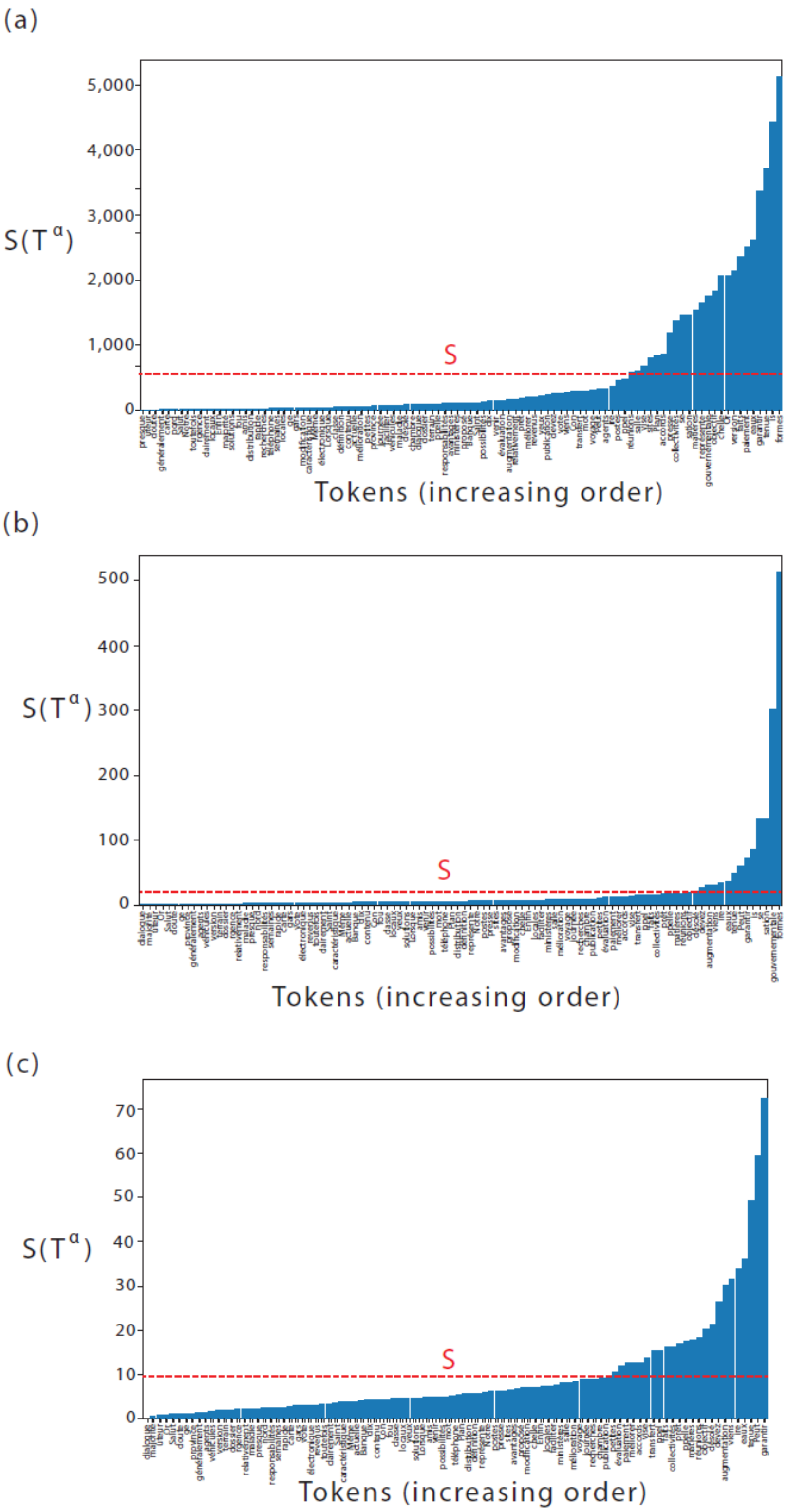

**Fig. 5.** Similar to Fig. 4, but with source sentences in French and generated sentences in English.

These results clearly demonstrate that mutual English-French translation using the MarianMT (Helsinki-NLP/opus-mt) system is asymmetric. This asymmetry likely arises from an intrinsic difference between the two languages rather than the translation model itself, as the mutual English-Hebrew translation by MarianMT (Helsinki-NLP/opus-mt) exhibits near symmetry in $S^{95}$ (Table 1).

| K lowest $S(T^{\alpha})$ | English to French | French to English | English to Hebrew | Hebrew to English |
|---|---|---|---|---|
| 100 | 29.5 | 20.7 | 8.0 | 17.5 |
| 95 | 3.6 | 9.5 | 5.7 | 6.3 |

**Table 1.** Mutual English-French and English-Hebrew translation entropies estimated by Marian (Helsinki-NLP/opus-mt). $S$ (first row) and $S^{95}$ (second row) calculated using $100$ pivot source tokens with thresholds as in Figs. 4 and 5.

### 2.3. Multiplicative translation degeneracy effect using two pivot tokens

The translation degeneracy examined for a single pivot source token, $SG(T^{\alpha})$, (Section 3.1) was extended to the case of two pivot source tokens, $SG(T^{\alpha}, T^{\beta})$. First, the individual degeneracies $SG(T^{\alpha})$ and $SG(T^{\beta})$ of a sentence containing tokens $T^{\alpha}$ and $T^{\beta}$ were calculated. Next $SG(T^{\alpha}, T^{\beta})$ was estimated by generating $|SG(T^{\alpha})| \cdot |SG(T^{\beta})|$ sentences—representing all combinations where the two tokens were replaced—and determining how many of these were identical to the source sentence. Owing to the high computational complexity of this procedure, it was performed for only about $100$ sentences, with detailed results illustrated for $13$ representative sentences (Fig. 6). The results indicate that

$$SG(T^{\alpha}, T^{\beta}) > 0.5 \cdot SG(T^{\alpha}) \cdot SG(T^{\beta}) \quad (6)$$

This finding indicated that the degeneracy for two tokens is approximately proportional to the product of their individual degeneracies, with a slight reduction factor. The reduction suggests that translation is not applied to each token independently; rather, the two tokens exhibit semantic correlation within the sentence.

| Source English sentence | $SG(T^1)$ | $SG(T^2)$ | $\frac{SG(T^1,T^2)}{SG(T^1)\cdot SG(T^2)}$ |
|---|---|---|---|
| These projects cover the full spectrum of physical, chemical and biological measurements and testing in a wide range of different areas relating to the needs of industry, food and agriculture, environmental monitoring, health and safety in the work place, health care, customs, the justice system and the protection of cultural heritage. | 245 | 9 | $\frac{1{,}724}{2{,}205}=0.78$ |
| You seemed very much in love, your arms full of wine and food | 86 | 26 | $\frac{1{,}132}{2{,}236}=0.51$ |
| You are the craziest man I've ever met. | 4 | 48 | $\frac{140}{192}=0.73$ |
| You are more in need of a night in Atlantic City than any man I've ever met. | 14 | 69 | $\frac{532}{966}=0.55$ |
| That is why we read, "If a brother or sister be naked, and destitute of daily food, And one of you say unto them, Depart in peace, be ye warmed and filled; notwithstanding ye give them not those things which are needful to the body; what doth it profit? Even so faith, if it hath not works, is dead, being alone." (JAMES 2:15-17). | 2 | 188 | $\frac{342}{376}=0.91$ |
| He tried to walk-on here, but, uh... didn't make the team. | 5 | 29 | $\frac{136}{145}=0.94$ |
| Michael, because you became a policeman, doesn't mean you have to eat like a dog. | 16 | 27 | $\frac{301}{432}=0.70$ |
| He had nothing to eat when he was a kid, but now he treats his dog like a gentleman. | 9 | 48 | $\frac{330}{432}=0.76$ |
| Do you always play these kiddie games.. ..or, are you interested in something else too? | 2 | 81 | $\frac{137}{162}=0.85$ |
| One kid tried to play another kid. | 2 | 53 | $\frac{97}{106}=0.92$ |
| I have another question, I was using the internet to translate a French children's book one word at a time and I become lost. I think I understand it now but I want to make sure. | 12 | 2 | $\frac{20}{24}=0.83$ |
| As the opponents want us to believe, the Holy Quran is not the product of the holy Prophets speculation and thinking, rather it is a revealed book in which every word is the word of the Almighty that was communicated to the Prophet through the process of revelation. | 14 | 6 | $\frac{64}{84}=0.76$ |
| Over 73% of British Columbia farms reporting a computer used it for word processing, over 72% reported using it for bookkeeping, while 70% of farms reported using the Internet. | 162 | 3 | $\frac{395}{486}=0.81$ |

**Fig. 6.** Multiplicative translation degeneracy effect is exemplified using 13 sentences, with translation degeneracies $SG(T^1)$ and $SG(T^2)$, respectively, and their combined multiplicative translation degeneracy $SG(T^1,T^2)$ (Eq. (6)).

### 2.4. Ranking three translation models using TE

The proposed method for measuring TE was extended to two additional models: the T5-Base (Google) translator[26] and the NLLB-200 (nllb-200-distilled-600M) (Facebook) translator[27]. Each model comprises $12$ encoder blocks and $12$ decoder blocks, with $\sim 223$ million and $\sim 615$ million parameters, respectively.

The English-French TE for all three translators was estimated using the same $100$ pivot source words, and following the same procedure as for MarianMT (Helsinki-NLP/opus-mt) (Figs. 4c and 5c). Based on the $S^{95}$ measure, the T5-Base (Google) translator achieved the lowest TE (Table 2, second row), although its overall $S$ (Table 2, first row) was the highest due to a few exceptionally large $S(T^{\alpha})$ values. Interestingly, MarianMT's TE was comparable to T5-Base (Google) and much lower than NLLB-200 (Facebook), despite MarianMT having only six encoder and six decoder blocks and much fewer parameters ($\sim 75$ million compared to NLLB-200 (Facebook) $\sim 600$ million). However, for French-English translation, MarianMT outperformed the other two translators (Table 3, second row). The results suggest that the simpler MarianMT architecture may be the most efficient for mutual English-French translation among the three. However, differences in training datasets and procedures among the models may also influence these outcomes.

The same ranking order among the three translators was obtained for $\beta_c = 9$ (Eq. (4)). For translation from English to French, the approximated entropies were $(1.5, 2.8, 1.1)$ for (MarianMT, NLLB-200, T5-Base), and for translation from French to English, they were $(2.8, 6.5, 3.9)$, respectively. These results confirm the robustness of the proposed entropic measure to the selected threshold.

| K lowest $S(T^{\alpha})$ | S | | |
|---|---|---|---|
| | MarianMT | NLLB-200 | T5-Base |
| 100 | 29.5 | 73.5 | 90.9 |
| 95 | 3.6 | 13.0 | 2.8 |

**Table 2.** Comparison of TE from English to French among three translators, $S$ (first row) and $S^{95}$ (second row), using $100$ pivot source tokens with thresholds as in Fig. 4.

| k lowest $S(T^{\alpha})$ | S | | |
|---|---|---|---|
| | MarianMT | NLLB-200 | T5-Base |
| 100 | 20.7 | 251.2 | 394.0 |
| 95 | 9.5 | 108.9 | 295.9 |

**Table 3.** Comparison of TE from French to English among three translators, $S$ (first row) and $S^{95}$ (second row), using $100$ pivot source tokens with thresholds as in Fig. 5.

### 2.5. TE without threshold

The results were obtained using threshold, $\beta_c = 5$ (Eq. (4)), where most of the numerous meaningless tokens, which replace the pivot token with very low probabilities, were excluded (Fig. 3b). Neverthless this type of translation noise exists and its effect on the above-mentioned reported results and trends (using $\beta_c = 0$) is exhibited in Tables 4-6. The translation noise significantly increases the TE, however, a comparison between Tables 4-6 and Tables 1-3, respectively, indicates similar trends and ranking between the three translator models.

Using the $S^{95}$ measure, asymmetry between mutual English-French TE was observed where TE from French to English increased by ~3 times in comparison to the TE from English to French, however, the mutual TE between English and Hebrew is much more symmetric (Table 4). The TE of MarianMT is lowered in comparison to NLLB-200 and T5-Base translators for both English to French (Table 5) and French to English (Table 6) translations. These trends are almost similar to those obtained with threshold (Tables 1-3), although sampling the tail of the distribution of tokens which replace the pivot token with very low probabilities is expected to require much more than $100$ pivot tokens.

| K lowest $S(T^{\alpha})$ | English to French | French to English | English to Hebrew | Hebrew to English |
|---|---|---|---|---|
| 100 | 386.6 | 553.4 | 67.6 | 204.6 |
| 95 | 116.1 | 379.9 | 35.5 | 49.1 |

**Table 4.** Mutual English-French and English-Hebrew translation entropies estimated by Marian (Helsinki-NLP/opus-mt). $S$ (first row) and $S^{95}$ (second row) calculated using $100$ pivot source tokens without threshold.

| K lowest $S(T^{\alpha})$ | S | | |
|---|---|---|---|
| | MarianMT | NLLB-200 | T5-Base |
| 100 | 386.6 | 1,909.5 | 584.2 |
| 95 | 116.1 | 1,374.3 | 258.6 |

**Table 5.** Comparison of TE from English to French among three translators, $S$ (first row) and $S^{95}$ (second row), using $100$ pivot source tokens without threshold.

| k lowest $S(T^{\alpha})$ | S | | |
|---|---|---|---|
| | MarianMT | NLLB-200 | T5-Base |
| 100 | 553.4 | 3,475.2 | 1,413.9 |
| 95 | 379.9 | 2,840.6 | 1,176.9 |

**Table 6.** Comparison of TE from French to English among three translators, $S$ (first row) and $S^{95}$ (second row), using $100$ pivot source tokens without threshold.

The TE measure without threshold ($\beta_c = 0$) enables to estimate the enhanced translation along the decoder blocks, independent of the selected threshold which might need to adjust along the decoder blocks. The underlying learning mechansim along the six decoder blocks of MarianMT (Helsinki-NLP/opus-mt) trained on Opus100 dataset was quantified using the following method[28, 29]. The weights of the first $m$ decoder transformer blocks were kept unchanged (frozen)[30, 31], and the $768$ outputs were fully connected to the output units. This fully connected layer was trained using the Opus100 dataset, as previously used to train the entire MarianMT. Finally, the TE entropy of the truncated MarianMT with only $m$ decoder blocks was estimated using $S^{95}$ (Table 7). Results indicate improved translation along the decoder blocks where TE decreases. The fully connected layer was trained for $50$ epochs, using a CosineAnnealingLR[32, 33] with a learning rate of $1e-4$, while all other AdamW hyperparameters were kept at their default values[34].

| Decoder blocks | 1 | 2 | 3 | 4 | 5 | 6 |
|---|---|---|---|---|---|---|
| $S^{95}$ | 10,712 | 6,114 | 3,295 | 908 | 147 | 116 |

**Table 7.** Estimation of English to French TE along the $6$ decoder blocks of the MarianMT (Helsinki-NLP/opus-mt) trained on Opus100 using $S^{95}$.

## 3. Discussion

Several metrics exist for quantifying translation quality[35-37], with the two most widely used being COMET[38] and BLEU[39]. The BLEU metric evaluates machine translation quality by measuring the model's output to a reference translation, focusing on word choice and structural similarity. However, BLEU often fails to recognize cases where different phrasing still conveys the same meaning. In contrast, COMET applies advanced neural models that understand linguistic context and semantics. It compares the source sentence, the reference translation, and the model's translation to estimate meaning similarity based on patterns learned from human evaluations. BLEU scores range from $1$ to $100$, while COMET scores fall between $0$ and $1$. In both cases, higher values indicate better translation quality[40]. The evaluation of the three translators (Table 8) indicates that Marian MT performed best for both English-to-French and French-to-English translation based on COMET and BLEU scores. The second-best translator, based on the COMET score, was NLLB-200 (Facebook). In contrast, using the proposed entropic measure, the best English-to-French translation was T5-Base (Google) (fine-tuned on $200{,}000$ Opus100 examples) (Table 8). For French-to-English translation, MarianMT (Helsinki-NLP/opus-mt) ranked highest, using the entropic measure aligning with both BLEU and COMET results. Notably, COMET score differences were relatively small compared with those of BLEU (Table 8) and the proposed entropic measure (Tables 2 and 3). Additionally, the non-fine-tuned version of T5-Base (Google) produced lower scores than its fine-tuned counterpart, despite fine-tuning being performed on only a small subset of the Opus100 dataset (Table 8).

The underlying learning mechansim along the decoder transformer blocks was quantitatively estimated using MarianMT (Helsinki-NLP/opus-mt) trained on Opus100. Results indicate improvement in translation along the decoder blocks where TE decreases (Table 7). Similar results were obtained for COMET and BLEU measures where their score increases along the decoder transformer blocks (not shown).

| Model | English to French | | French to English | |
|---|---|---|---|---|
| | BLEU | COMET | BLEU | COMET |
| MarianMT (Helsinki-NLP/opus-mt) | 38.83 | 0.8026 | 39.82 | 0.8223 |
| NLLB-200 (Facebook) | 33.27 | 0.798 | 34.38 | 0.8037 |
| T5-Base (Google) (fine-tuned on 200,000 examples of Opus100) | 37.08 | 0.7763 | 28.19 | 0.7299 |
| T5-Base (Google) | 36.79 | 0.7759 | 5.63 | 0.6979 |

**Table 8.** BLEU and COMET scores for English-to-French and French-to-English translations. The last row (gray) presents the results for T5-Base (Google) before fine-tuning.

The proposed entropic measure for estimating translation quality has some limitations. Multiple sets of probabilities $P_i$, can yield the same entropy per pivot token (Eq. (2)), though they may differ in interpretation. One limiting case is that zero entropy can occur regardless of the number of tokens with $P_i = 1$, representing a pivot token with several strong-match tokens. Another limiting case occurs when one pivot token has few high $P_i$ values (e.g., $P_i = 9$) and another has many low values (e.g. $P_i = 1/24$); both can yield the same entropy but represent different linguistic realities. However, the quantities per pivot token $|SG(T^1)|$ and $N_{Av}(T^1)$, can help distinguish between these cases. Therefore, it is suggested refining the interpretation of entropy per pivot token and $S$ using these additional quantities. Another limitation arises from computational constraints: TE and $S$ were estimated using only $100$ pivot tokens and $30$ sentences for each. Larger datasets are needed to obtain more robust statistics for TE estimation.

The proposed TE method assumes that for a given dataset lower entropy corresponds to a better translator. However, as each language naturally contains synonyms—tokens that serve as strong or weak matches to a pivot token—a zero-entropy limit cannot realistically represent a language. As the finite minimal achievable entropy of any language remains unknown, minimizing TE only moves translation quality toward this unknown lower bound.

An asymmetry in the mutual TE between English and French was observed using the MarianMT translator, whereas the English-Hebrew translation appeared nearly symmetric. One must analyze the TE across multiple translators and languages to determine whether such asymmetry arises from linguistic structure or from the properties of the translation model. If language structure is indeed responsible for asymmetry, it would be valuable to cluster languages according to whether their mutual TE is symmetric or asymmetric and to examine how many such clusters exist.

The advanced performance of MarianMT ($\sim 75$ million weights) compared with the NLLB-200 (Facebook) ($\sim 615$ million weights) and T5-Base (Google) ($\sim 215$ million weights) translators (Tables 8 and 3) indicates that larger models do not necessarily produce better translations. However, the NLLB-200 (Facebook) model is a multilingual translator, designed for a broader task than MarianMT. This observation raises the questions about the optimal balance between translation quality and the size and structure of deep translational architectures. Currently, the examined translators operate in a regime where the number of model weights far exceeds the number of tokens, suggesting that smaller architectures may achieve comparable performance. Moreover, factors such as optimization of the training dataset, sentence types, token frequency distributions, and tokenizer quality are expected to influence translation performance, warranting further research.

Finally, the entropy per pivot token fluctuates, with a small fraction of tokens exhibiting high entropy while most remain low (Fig. 4). Developing a pretraining procedure to suppress the entropy of high-entropy pivot tokens represents an important direction for future research.

## Acknowledgements

We thank Yarden Tzach for helpful comments and discussions.